\documentclass[sigconf]{acmart}

\usepackage{booktabs} 

\setcopyright{rightsretained}

\acmConference[CoDS-COMAD'19]{The ACM India Joint International Conference on Data Science \&
Management of Data}{January 3-5, 2019}{Kolkata, India}
\copyrightyear{2019}

\begin{document}
\title[Extractive summarization using keyword frequency and entity graphs]{Effective extractive summarization using frequency-filtered entity relationship graphs}

\author{Archit Sakhadeo}
\affiliation{%
  \institution{Department of Computer Engineering, Pune Institute of Computer Technology, University of Pune}
  \city{Pune}
  \country{India}
}
\email{architsakhadeo@gmail.com}

\author{Nisheeth Srivastava}
\affiliation{%
  \institution{Department of Computer Science and Engineering, Indian Institute of Technology Kanpur}
  \city{Kanpur}
  \country{India}
}
\email{nsrivast@iitk.ac.in}

\renewcommand{\shortauthors}{Archit Sakhadeo \& Nisheeth Srivastava}

\begin{abstract}
Word frequency-based methods for extractive summarization are easy to implement and yield reasonable results across languages. However, they have significant limitations - they ignore the role of context, they offer uneven coverage of topics in a document, and sometimes are disjointed and hard to read. We use a simple premise from linguistic typology - that English sentences are complete descriptors of potential interactions between entities, usually in the order subject-verb-object - to address a subset of these difficulties. We have developed a hybrid model of extractive summarization that combines word-frequency based keyword identification with information from automatically generated entity relationship graphs to select sentences for summaries. Comparative evaluation with word-frequency and topic word-based methods shows that the proposed method is competitive by conventional ROUGE standards, and yields moderately more informative summaries on average, as assessed by a large panel (N=94) of human raters.
\end{abstract}

%
%



\maketitle

\section{Introduction}
The use of statistical methods for automatic summarization of text dates back as far as 1958, when Luhn~\cite{luhn} suggested that the author of any document will use certain words related to the central topic more often than words not related to the topic. Thus words with more frequency had higher importance, in that they would convey information related to the most important topics present in the document. Furthermore, the significance of a sentence was determined by the relative position of the high frequency words within a sentence. 

Statistical methods have since evolved considerably in sophistication. In Latent Semantic Analysis (LSA) based summarization~\cite{lsa}, a term-sentence matrix is formed and Singular Value Decomposition (SVD) is applied on this matrix. 
The k\textsuperscript{th} singular vector represents the k\textsuperscript{th} important concept. The sentences that best represent the corresponding important singular vectors are chosen in the summary. 

Another statistical approach, TextRank~\cite{textrank}, is a graph-based model that uses a PageRank~\cite{pagerank} based ranking algorithm to assign relevance scores to sentences. For sentence extraction, sentences are considered as vertices and similarity between sentences is used to obtain a weighted graph. The ranking algorithm is run on this graph and top sentences with higher scores are selected.

Yet another popular method, LexRank~\cite{lexrank}, uses graph-based centrality scoring of sentences to define sentence salience. It assumes that sentences that are similar to many other sentences are more central to the topic. On creating a similarity graph of sentences, centrality is computed through various ways. Sentences are ranked using a PageRank based ranking algorithm. It also implements cross sentence informational subsumption to avoid selecting very similar sentences in the summary as similar sentences lead to smaller coverage of text.

Complementing such purely statistical approaches to summarization, multiple researchers have pointed to the potential value of incorporating semantic information also. An early suggestion along such lines was made by Edmundson~\cite{edmundson}, who proposed using features like title, heading, position of sentence in the text, first sentences of paragraphs or last concluding sentence, etc., and cue words and phrases like `best', `important', `results' to score sentences. While statistical methods are much more principled, in practice, heuristic methods such as Edmundson's proposal yield nearly equivalent results in many important applications even today. 

More principled attempts to incorporate semantic knowledge into summarization algorithms have been developed by leveraging pre-existing ontologies of object relationships. For example, by mapping sentences to concepts in an ontology using multi-label classification, Hennig et al.~\cite{hennig2008ontology} used a semantic representation of sentences to select important ones using graph-theoretic properties. Similarly, Baralis et al.~\cite{baralis2013multi}  have proposed a summarizer that leverages a Wikipedia-based large open-source ontology to identify important concepts in documents, so that sentences containing them may be selectively extracted.

The summarization method we propose in this paper follows in this second vein of research, trying to leverage extra-statistical information to improve text summarization. Our point of departure from the previous semantically informed techniques for extractive summarization is that our method acquires entity-relationships using linguistic structure instead of pre-existing ontologies. Comparative evaluation with existing prominent models shows competitive performance on ROUGE metrics on two DUC datasets. Evaluation by a large panel of human raters suggests that summaries generated using the proposed method rate highly on both informativeness and coherence, and are more likely to be misclassified as generated by humans than alternative approaches.

\section{A hybrid graph and keyword-based summarizer}

The starting point for our project was the observation that keyword-based and graph-based approaches for automatic summarization have complementary strengths and weaknesses. Keyword frequency based methods provide superior precision in identifying the key topics and its associated content in a document. But they do this at the expense of omitting contextual detail that is associated conceptually, but not syntactically, with the central topic in the text. On the other hand, graph-based and topic modeling based methods provide superior coverage of such related topics, but frequently have difficulty adjusting the relative weights for each of these topics in the summary, resulting in disjointed and incoherent summaries. 

The underlying trade-off across these methods is the level of granularity at which information about what is important in the text is represented. Keyword-based approaches store the relative significance of topics in frequency counts, but lack the ability to identify distinct topics in the text. Graph-based methods possess the latter ability, but cannot assess the relative weight of these topics. In this paper, we design a simple algorithm that attempts to store both aspects of information - topics and their relative weights. Further, it does so in an unsupervised, language-agnostic manner such that the algorithm can be applied to any language, given the existence of an entity tokenizer and a co-reference resolution tool.

\begin{figure}[h]
     \centering
     \includegraphics[width=0.45\textwidth]{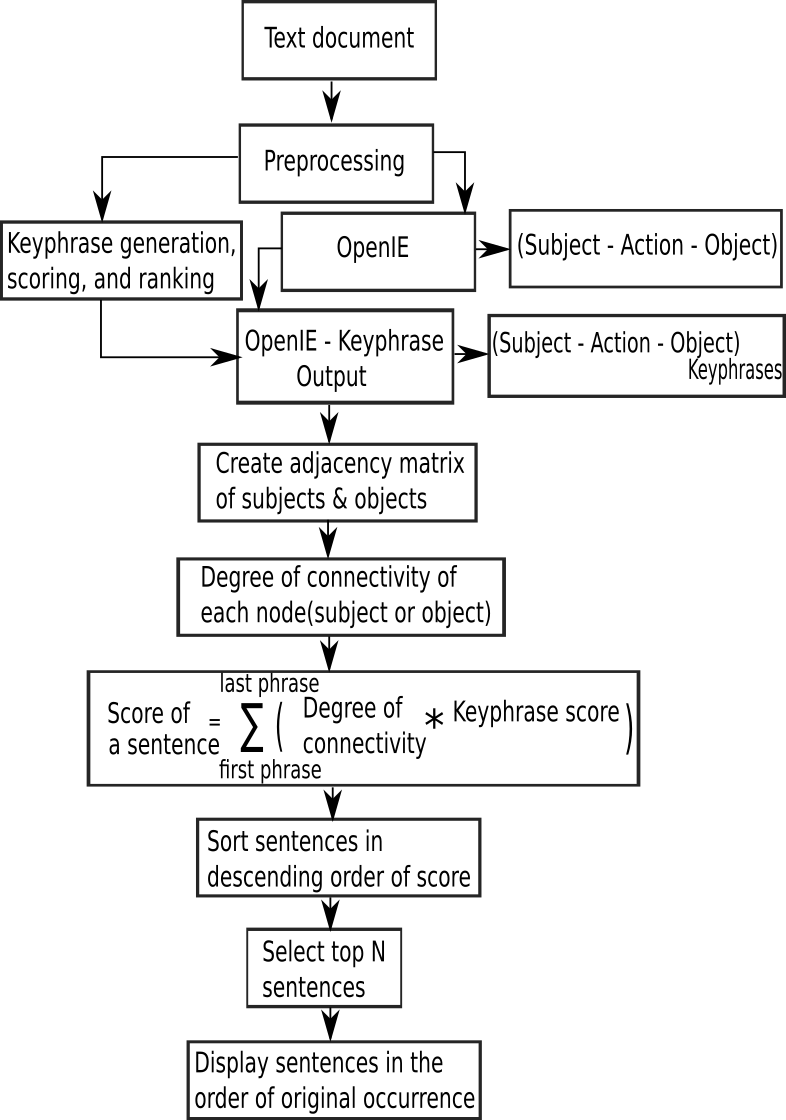}
     \caption{A joint word-frequency and entity relationship based extractive summarization method}
     \label{fig:flowdiagram}
 \end{figure}

The basic concept of our approach is that we extract keywords and entity relationships independently using existing tools, but then take the intersection of these two output streams using a novel scoring strategy that focuses attention on keywords that semantically represent important entities in the original text. Figure \ref{fig:flowdiagram} schematically describes our summarization method. We further present a step-wise description below.

{\bf Preprocessing}: The original text documents were preprocessed to manage non-linguistic markers masquerading as sentence completion symbols. This was particularly important, since our approach relies heavily on identifying and further analyzing individual sentences in the document. We check for the following situations as they need to be treated differently than full stops: initials like J. K. Rowling, decimal points like the floating point number 1.24, designation symbols like Dr., Mr., Mrs., Ms., etc. Sentence tokenization was done after dealing with the above situations. Consecutively occurring proper nouns (Name + Surname) are treated as one entity.

{\bf Keyword and keyphrase extraction}: The text in the document is then split on multiple punctuation marks, stopwords, and verbs as our primary focus is on selecting keywords which are entities. The resulting text is further split on spaces. Each such word obtained is called as a keyword. If the keywords occur consecutively within the original sentence, they form a keyphrase. For example, `Indian aeroplane is best of its kind ; thus gaining world ranking 3 in terms of safety' would be converted to `Indian aeroplane | best | kind | world ranking 3 | terms | safety'  where | is a delimiter. Here, `Indian', `aeroplane', `safety', etc. are keywords but `Indian aeroplane' is a keyphrase. All stopwords are removed from the text as they are high frequency words that do not add important information to sentences, and instead would incorrectly rank higher in terms of score in the keyword list.

Following earlier word-frequency based approaches like RAKE~\cite{rake}, scoring of the keywords was done based on their keyword frequencies and keyword degrees with the score given by equation \ref{score}
\begin{equation}
    \label{score}
    \text{score} = \frac{\text{Degree(keyword)}}{\text{Frequency(keyword)}} 
\end{equation}

Keyword frequencies are calculated by finding the number of times a particular keyword occurs within the document. Keyword degree in a sentence is calculated by counting the number of keywords that co-occur with that particular keyword within a sentence. Thus, the total degree of a keyword over a document is the summation of all such keyword degrees in sentences which have an instance of that particular keyword. Scores are calculated for each keyphrase based on sum of scores of respective keywords. This score prioritizes high frequency words and phrases that occur across multiple sentences, and thus have high coverage of text and association with other words. 

{\bf Entity-relationship extraction}: We use the linguistic typology of the English language, viz. its SVO structure~\cite{tomlin1986basic} to learn (Subject $\xrightarrow{}$ Action $\xrightarrow{}$ Object) structures from English sentences using grammar parsing, and then extract entity relationships from these subject-object pairs. We operationalize this inference using the Open Information Extractor (OpenIE)~\cite{angeli2015leveraging} which splits sentences into maximally shortened entailed clauses and outputs data structures in the form [ [S1,A1,O1], [S2,A2,O2], [S3,A3,O3], ... , [Sn,An,On] ] \newline where each [Sm,Am,Om] is a triple from a sentence where Sm, Am and Om correspond to subject, action and object respectively.

OpenIE output in its base form contains a lot of noise i.e. redundant triples. We processed this output to merge or delete redundant or incorrectly inferred triples to get them in an acceptable form. We also used OpenIE's native coreference resolution tool to resolve some grammatical redundancies, e.g. pronoun use, automatically. 

This did not completely resolve the redundancies so, in addition, we made use of anaphoric reference by replacing the pronoun subjects in a given sentence by the noun subjects in the previous sentence by identifying the voice of the sentences. The pronoun may even refer to the object of the previous sentence. However, in practice we found that it was likely to refer the subject.

\begin{figure}[h]
     \centering
     \includegraphics[scale=0.40]{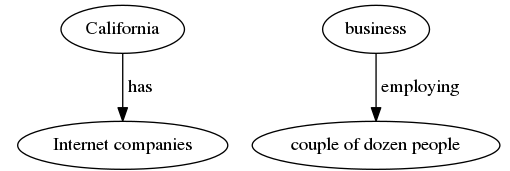}
     \caption{Subject-Action-Object graphs}
     \label{fig:graphs}
 \end{figure}

Here, we take the intersection of the extracted keywords and the entity relationships. Subjects from the processed triples that are also present in the extracted keyphrases are now selected. These subject-keyphrases combination have more significance than individual keyphrases or subjects as they appear prominent by both statistical and semantic criteria. Triples corresponding to these filtered subjects are taken into consideration for further processing. We build graphs like the ones in Figure \ref{fig:graphs}.

{\bf Scoring of words and sentences}: Subjects and objects from these filtered triples are separately considered as individual nodes within a graph. An adjacency matrix of all the nodes is formed. The matrix will thus represent a directed graph with connectivity scores of each node equal to the number of outgoing edges from that node. This gives us the connectivity of each node from the matrix.

These outgoing edges, locally within one sentence, represent only subject to object connectivity. But for the graph obtained from all sentences, it could also represent subject to subject, object to subject, or object to object mappings since one sentence's subject can be another's object. We then assign scores to each sentence based on the presence of the nodes and their connectivity i.e. the score of a sentence being equivalent to the sum of the connectivity scores of each node present in it. Alternative scoring method is discussed below in equation \ref{finalscore} in Section \ref{sec:alternates}. 

{\bf Selection of sentences}: The sentences are sorted in the descending order of the calculated scores. Top N\% of these sorted sentences are selected to be included in the summary. We then display them as per their order of occurrence in the original document. This forms our summary for the original input.

\section{Evaluation}
Baseline summarizers denote the five external summarizers that we have chosen to evaluate and compare against our approaches. System summarizers denote all the eight summarizers, external and our approaches included. We used the Document Understanding Conference (DUC) datasets from the years 2001~\cite{duc2001} and 2002~\cite{duc2002} for evaluation. The datasets consist of:

1) Original text documents - These are the documents which are to be summarized. We use these documents as inputs to the system summarizers and compare their output summaries.

2) Gold summaries - These documents are the human-generated abstractive summaries provided by DUC. We use these gold summaries as our reference to check how human-like the system summaries are. The system summaries of each document are independently compared with the corresponding gold summary and are evaluated to get the recall, precision, and F score.

The datasets had extracts (original text documents) and abstracts (gold summaries). It had multiple extracts on a topic and corresponding abstracts for each extract. 599 documents from the 2001 dataset and 536 documents from the 2002 dataset i.e. 1135 text documents were used for processing and evaluation.

\subsection{Flavors of our approach}
\label{sec:alternates}
We empirically tested three different variants of the baseline model described above, denoted Summ-W, Summ-NW, Summ-NW-K*S, for reasons that become clear below. 

{\bf Summ-W (Winnowing Method)}: Here we tried to first convert the text to a shorter intermediate form and then to the final form. We captured the top 3-4 most connected nodes and filtered the sentences containing any of these important nodes to get the intermediate stage. This intermediate stage text was about 60\%-70\% of the total document size. Thus, the procedure was to reduce the document to a 60\%-70\% sized text and consider this intermediate stage text as our new input. All the steps mentioned in Figure \ref{fig:flowdiagram} are performed on this new smaller input to further generate the final summary. Our hypothesis was that by doing this, we will keep the unimportant sentences away and then run our system on the intermediate stage text to get better results. This approach is represented by Summ-W. It turned out that by finding the intermediate stage, some sentences which did not contain the top 3-4 nodes but which still captured more information had been excluded. 

{\bf Summ-NW (Non-winnowing Method)}: Later, we introduced the Summ-NW summarizer. In this, the intermediate stage was removed and the system was tested again with summary being calculated directly from the original document using all the steps mentioned in Figure \ref{fig:flowdiagram}. The system captured the sentences which would otherwise have been excluded in the Winnowing method. The results were better in this case. Though the precision decreased by an average of 1.84\%, there was an average increase of 4.04\% in the recall and led to a slight increase in the value of F score. Even the quality of summaries in this approach were slightly better than Summ-W.

{\bf Summ-NW-K*S (Non-winnowing Method)}: In both the earlier methods, the sentence score was calculated using the connectivity of nodes i.e. by counting the outgoing edges of the nodes. The distinction between important words can further be enhanced by using the connectivity scores along with the keyphrase scores as a metric to rank words, and in turn sentences. That is by considering equation~\ref{finalscore}, we would get a better separation between words, and in turn sentences, as it would give higher scores to words that are prominent by both statistical and semantic criteria. Here, every keyphrase includes one or more words.\begin{equation}
    \label{finalscore}
    \text{final score} = {\text{connectivity score}}\times{\text{keyphrase score}}
\end{equation}

Using equation~\ref{finalscore} we prioritize nodes that have higher connectivity and higher keyphrase score. Sentences consisting of such highly scored phrases are captured in the summary. Thus, slight modifications were made in the selection of summary sentences in the Summ-NW method and the resulting summarizer is labelled as Summ-NW-K*S. It led to minor improvements in the precision, F score, and also the quality of the summaries.



\subsection{ROUGE evaluation}

We used ROUGE~\cite{rougepaper}, the {\em de facto} evaluation metric for extractive summarization methods, to calculate the recall, precision and F score of the system summaries with respect to their corresponding gold summaries (reference summaries here). All system summaries had the same number of sentences for a given percentage length.

The reference summaries are the texts that are used as a standard to compare the system summaries against. We use gold summaries as our reference here because we aim to check how the system summaries perform with respect to their corresponding human-generated summaries.

Values of recall, precision, and F score vary between 0 and 1. Recall measures how much text from the reference summary is captured in the system summary and precision measures how much excess text is present in the system summary that is not in the reference summary. We need more relevant information to be captured (high recall) and less excess information (high precision). A combined effect is taken into consideration by taking the harmonic mean of recall and precision. This is called the F score. Higher the recall, precision and F scores of a system summary with respect to the reference gold summaries, statistically closer is the system summary to the reference human-generated summary.

\begin{figure}[h]
     \centering
     \includegraphics[scale=0.35]{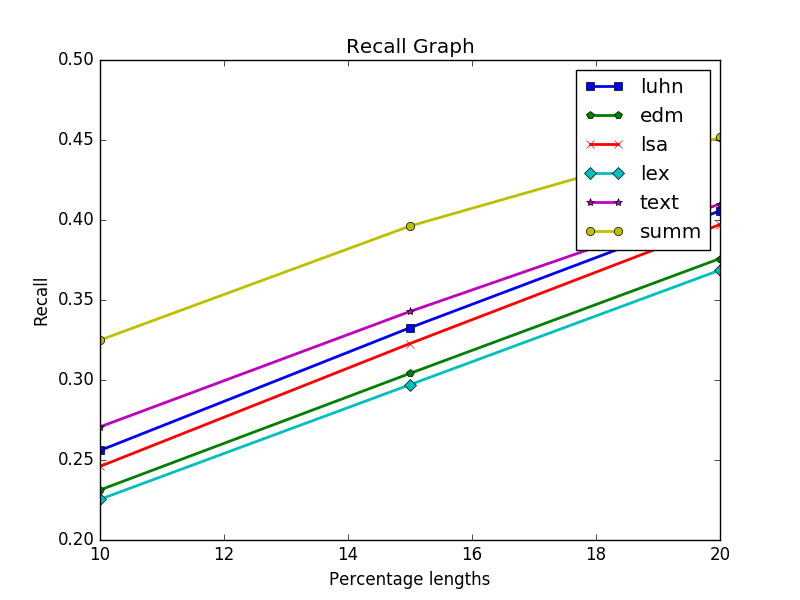}
     \caption{Recall vs summary length graph}
     \label{fig:recall}
 \end{figure}

\begin{figure}[h]
     \centering
     \includegraphics[scale=0.35]{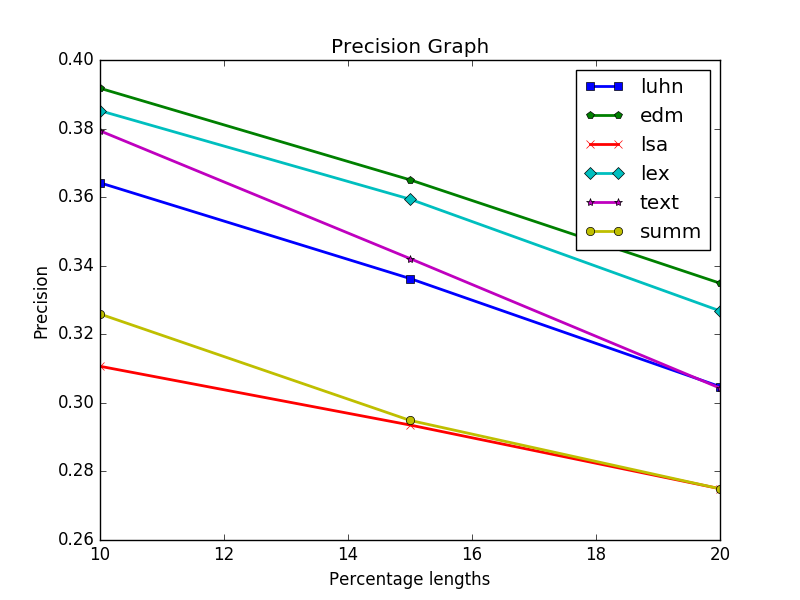}
     \caption{Precision vs summary length graph}
     \label{fig:precision}
 \end{figure}

\begin{figure}[h]
     \centering
     \includegraphics[scale=0.35]{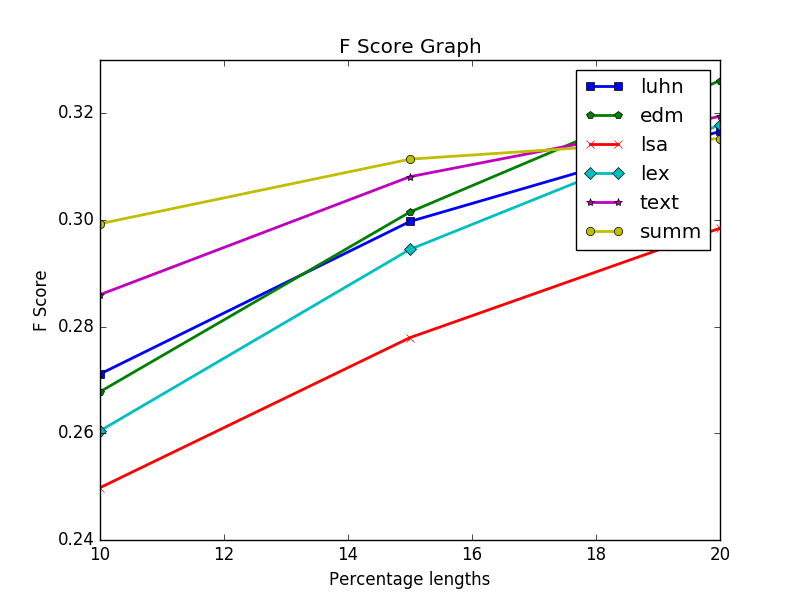}
     \caption{F score vs summary length graph}
     \label{fig:fscore}
 \end{figure}

For both datasets, we recorded the results using three different scenarios: system summary lengths - 1) equal, 2) shorter, 3) longer - than their corresponding reference gold summaries which were about 15\% of the original text documents on average. For every system summarizer, we produced summaries of three different lengths for the same text document, and using ROUGE we compared them with the gold summaries (constant length). The three lengths were 10\%, 15\%, and 20\% of the text document. ROUGE results corresponding to the varying length summaries give insights into the features of the summaries as explained in Section~\ref{sec:interpretation}. Refer Figure~\ref{fig:recall}, Figure~\ref{fig:precision}, and Figure~\ref{fig:fscore} for the correlation between ROUGE scores and summary lengths of Summ-NW-K*S (represented by summ) and the baseline summarizers (represented by luhn, edm, lsa, lex, text). From the three flavors of our summarizer, we used Summ-NW-K*S to represent our approach.

We used the Python sumy package~\cite{sumy} implementations of the baseline summarizers for comparison. Here the gold summaries' average length for DUC 2001 dataset is about 11\% of the text documents, and for DUC 2002 dataset it is about 18\%. Hence, results corresponding to 10\% lengths were selected for DUC 2001 dataset and average results corresponding to 15\% and 20\% were selected for DUC 2002 dataset. These scores were further weight-averaged based on the number of documents in the two datasets to give the combined results. Tables \ref{recall}, \ref{precision}, and \ref{fscore} collate these results.
\begin{table}[h] 
\begin{center}
\begin{tabular}{|l|rl|rl|rl|}
\hline   Recall &  Weighted &   Rank & DUC & DUC\\ 
& Average & &2001 &2002 \\ \hline
Luhn		&	0.3095 & 5   & 0.2628 & 0.3617\\
Edmundson	&	0.2803 & 7   & 0.2349 & 0.3312\\
LSA 		&	0.2983 & 6   & 0.2476 & 0.3550\\
LexRank		&   0.2777 & 8   & 0.2342 & 0.3265\\
TextRank	&	0.3204 & 4   & 0.2775 & 0.3685\\
Summ-W		&  \bf 0.3586 & \bf 3 & \bf 0.3284 & \bf 0.3924  \\
Summ-NW		&  \bf  0.3731 &  \bf 1 & \bf 0.3383 & \bf 0.4121  \\
Summ-NW-K*S	&  \bf 0.3715 &  \bf 2   & \bf 0.3376 & \bf 0.4095 \\
\hline
\end{tabular}
\end{center}
\caption{\label{recall} Average recall results - ROUGE 1, without stopwords }
\end{table}
\begin{table}[h]
\begin{center}
\begin{tabular}{|l|rl|rl|rl|}
\hline   Precision &  Weighted &   Rank & DUC & DUC\\ 
& Average & & 2001& 2002\\ \hline
Luhn		&	0.3431 & 4  & 0.3171 & 0.3723  \\
Edmundson	&	0.3733 & 1  & 0.3434 & 0.4070  \\
LSA 		&	0.2970 & 8  & 0.2651 & 0.3328   \\
LexRank		&   0.3700 & 2  & 0.3455 & 0.3974  \\
TextRank	&	0.3534 & 3  & 0.3332 & 0.3761  \\
Summ-W		& \bf  0.3082 & \bf 5 & \bf 0.2743 & \bf  0.3461  \\
Summ-NW		& \bf  0.3025 & \bf 7 & \bf 0.2685 &  \bf 0.3406 \\
Summ-NW-K*S	& \bf  0.3039 & \bf 6  & \bf 0.2700 & \bf 0.3418\\
\hline
\end{tabular}
\end{center}
\caption{\label{precision} Average precision results - ROUGE 1, without stopwords }
\end{table}
\begin{table}[h]
\begin{center}
\begin{tabular}{|l|rl|rl|rl|}
\hline   F Score &  Weighted &   Rank & DUC & DUC\\ 
& Average & & 2001 & 2002 \\ \hline
Luhn		&	0.2928 & 6 & 0.2562 & 0.3339 \\
Edmundson	&	0.2931 & 5  & 0.2541 & 0.3368 \\
LSA			&   0.2709 & 8 & 0.2313 & 0.3152 \\
LexRank		&   0.2894 & 7 & 0.2540 & 0.3291 \\
TextRank	&	0.3041 & 4 & 0.2716 & 0.3405 \\
Summ-W		& \bf 0.3078 & \bf 3  & \bf 0.2738 &\bf 0.3460\\ 
Summ-NW		& \bf 0.3101 & \bf 2  & \bf 0.2744 &\bf 0.3500\\ 
Summ-NW-K*S	& \bf 0.3107 & \bf 1 & \bf 0.2755 &\bf 0.3501\\
\hline
\end{tabular}
\end{center}
\caption{\label{fscore} Average F score results - ROUGE 1, without stopwords }
\end{table}


\subsection{Interpretation}
\label{sec:interpretation}
Our method shows substantial improvements over these baselines in terms of the composite F measure. We also find that the recall of our approaches is high, whereas the precision is low. 

Our method captures the highest recall for all the lengths. It can be observed that we capture more recall for even shorter summaries (10\%) than many other longer baseline summaries (15\%). Thus, we always capture more relevant information than the baseline summaries, even when our summaries are shorter. All system summaries had the same length (number of sentences) and thus the recall score is high not because of more number of sentences leading to the possibility of more relevant text for our method.

On the other hand, the precision of our method seems to be low. This suggests that the amount of text that is unwanted and is outside the scope of the reference gold summary is more in our summaries than other baseline summaries. This occurs because our method focuses on selecting sentences with higher scores to increase recall, thereby privileging sentences that are longer in length (more words). More keywords yield higher sentence scores, but they also mean more excess words, and in turn less precision. An alternative scoring method to work around this limitation is discussed further below using equation   \ref{ratio} in Section \ref{sec:discussion}. In spite of focusing on maximizing recall, thereby, tending to capture more words, ours summaries were shorter in length (number of words) than Luhn and TextRank's summaries. So, more number of words is certainly not the only and also not the most important factor in determining high recall. High recall inherently depends on the quality of sentences captured as well.

Our method has a higher F score even for our shorter summaries (10\% length) than many longer baseline summaries (15\% length). F score of our approach is higher than baseline summaries till 15\% length (reference summary length) after which it increases very slowly. While for all baseline summaries, F score increase after 15\% length is steeper than our F score increase. This suggests that the positive contribution our summaries have towards F score has decreased from 15\% to 20\% than from 10\% to 15\%, while for the baselines summaries the contribution has relatively increased. This effect is brought about by the additional sentences from 20\% that were not present in 15\%. The additional sentences at 20\% have contributed more for the baseline summaries than for our approaches. Additional sentences come with relevant information as well as excess words. For an ideal summarizer that ranks sentences in the decreasing order of the relevant information content, addition of new sentences to the system summary should lead to decrease in the new relevant information added and increase in new excess words added than the previous sentence. This should ideally lead to a decrease in the added F score value due to this combined effect of less recall and less precision of additional sentences. The effect is significant especially when system summaries are longer than reference summaries. 

The fact that the additional sentences from 15\% to 20\% have contributed more towards the F score of baseline summaries than our summaries, suggests that they have added more relevant information and less excess words than our sentences at 20\%. This information given that other summaries have low recall and F score while our summaries have high recall and F score for smaller summary lengths like 10\%, suggests that their additional sentences capture relevant information that they should have captured in their earlier highly ranked sentences. Thus, the relevant information comes in much later in their summaries. Unlike other summarizers, our approach includes all relevant information in the earlier highly ranked sentences as suggested by our high recall and F scores in shorter summaries.

Our summaries were approximately as long (number of words) as LSA's summaries. Even if number of words is considered as a metric of length instead of number of sentences, comparison between our approaches and LSA gives insights into how our approaches still fare better. Unlike LSA, our recall is high. Even the precision is slightly high. Interestingly, our approaches rank 1,2,3 in F score while LSA ranks 8 for the same summary size.

To summarize, our method provides the highest recall (always) and F score and does so for summary lengths shorter than other baseline summarizers. Our method also captures more relevant information earlier in the top few important sentences and it thus proves that we rank our sentences correctly and in a better way based on the relevant information they contain. This ensures maximum information gain in limited sentences. We would get similar results even if the length of gold summaries is increased, as long as our summaries are approximately equal in length to the gold summaries, as we would then add sentences that capture the next best relevant information. 
The total captured relevant information will always be higher than baseline summaries, leading to a higher recall and F score for system summary lengths less than or equal to reference summaries as it has been empirically observed.
\subsection{Human evaluation}
We also employed a direct human evaluation to see how our approach fares against baseline summarizers. All eight system summaries were not chosen as it would have led to large cognitive burden on participants and, in turn, insincere responses. We restricted the options to three summaries and chose TextRank's and Edmundson's summaries to compare against Summ-NW-K*S as they ranked overall better than other baseline summaries in terms of recall, precision and F score. Summ-NW-K*S was chosen from among our other two approaches as this was based on slightly better metrics to rank the sentences. 

The evaluation consisted of a form to be filled by the participants. 
Each form included an original text document and we presented three corresponding summaries. These three summaries were the outputs of Edmundson's, TextRank and Summ-NW-K*S. The summaries were randomly ordered and not labeled by the summarizers' names to get unbiased responses. Every participant got access to a different form and thus, to different texts. 

We asked the participants to rank the summaries independently based on three parameters:\\
1) Informativeness: How much relevant information does the summary capture?\\
2) Coherence: How organized and readable is the summary?\\
3) Opinion about the source: Has the summary been generated by a human by manually picking up important sentences or by an artificially intelligent system (AI)?

Informativeness and coherence ranged on a scale of 1-10 and participants were required to choose one score in this range for each of the parameters independently. The third parameter required participants to select the source as either human-generated or AI-generated. The thing to note here is that no summary was generated by a human. All were generated by some algorithm be it TextRank, Edmundson's or Summ-NW-K*S. The participants were unaware of this and were under the assumption that any or all of them could be human-generated or AI-generated. 

A total of 94 participants took the survey. The results for informativeness and coherence are aggregated on a scale of 10 while the opinion about the summary source is displayed as a measure of the number of people supporting either opinion. Refer Table \ref{informative}, Table \ref{coherence}, and Table \ref{opinion} for the results. Standard deviation (SD) is mentioned in the brackets along with the average scores. 

\begin{table}[h]
\begin{center}
\begin{tabular}{|l|l|}
\hline   &  Informativeness  \\ \hline
Edmundson	& 	6.48 / 10 (SD = 2.08) \\
TextRank	& 	7.06 / 10 (SD = 1.87)\\
Summ-NW-K*S	& {\bf 7.22 / 10} (SD = 1.92) \\
\hline
\end{tabular}
\end{center}
\caption{\label{informative} Average Informativeness scores }
\end{table}
\begin{table}[h]
\begin{center}
\begin{tabular}{|l|l|}
\hline   &  Coherence  \\ \hline
Edmundson	& 7.09 / 10	(SD = 1.87) \\
TextRank	& 7.47 / 10 (SD = 1.72)\\
Summ-NW-K*S	& {\bf 7.47 / 10} (SD = 1.72)\\
\hline
\end{tabular}
\end{center}
\caption{\label{coherence} Average Coherence scores }
\end{table}
\begin{table}[h]
\begin{center}
\begin{tabular}{|l|r|l|}
\hline   &  Human & AI  \\ \hline
Edmundson	& 34 & 60\\
TextRank	& 48 & 46 \\
Summ-NW-K*S	& \bf 50 & \bf 44\\
\hline
\end{tabular}
\end{center}
\caption{\label{opinion} Total Human vs AI counts}
\end{table}

 \begin{figure*}[h]
    \centering
     \includegraphics[width = 0.95\textwidth]{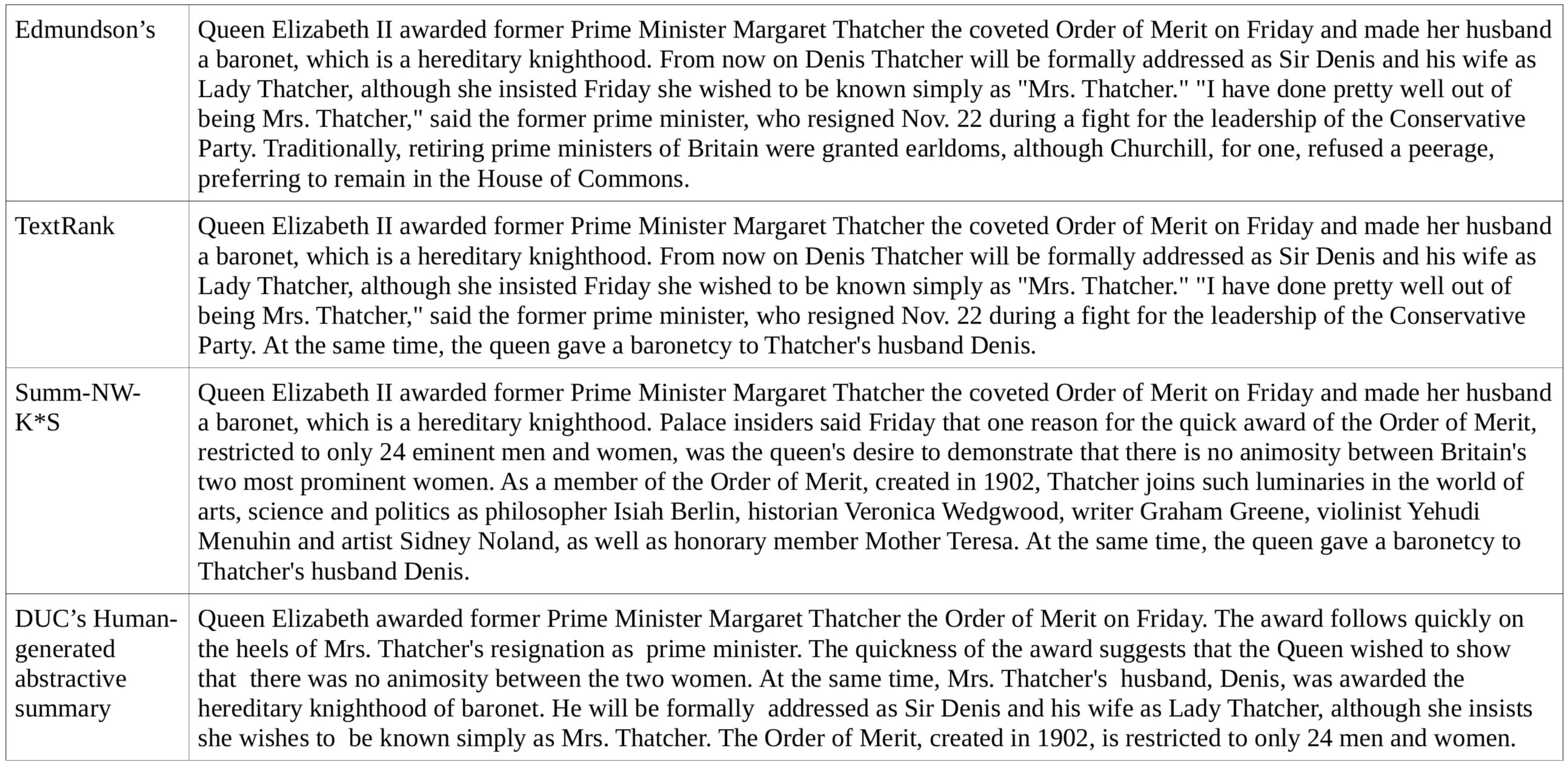} 
     \caption{\label{fig:comparison} Sample summaries from different summarization techniques applied to the same source document}
\end{figure*}

Table \ref{informative}, Table \ref{coherence}, and Table \ref{opinion} clearly show that our summaries ranked the best in all the three qualitative metrics. They were most informative and coherent, and participants misjudged our summaries to be generated by a human more than Edmundson's or TextRank's. Finally, to concretely illustrate the value of our new approach, we present a comparison of three sample summaries drawing upon the same source text in Figure \ref{fig:comparison}. Edmundson's summary inherently captures the first few lines of the text and then the concluding line, and thus lacks informativeness as it is focused in one area of the text. Similarly, TextRank happens to pick the first few lines in this example. In addition to the information covered in the first two summaries, Summ-NW-K*S captures other information regarding the absence of animosity, names of luminaries who have received the award, and rarity of the award. The summary by Summ-NW-K*S spans the whole text, captures more information and yet is coherent. Both the human-generated summary and our method's summary contain contextual information, e.g. rarity of the award, absence of animosity. Purely statistical and purely heuristic summarization methods omit such details, since they are not statistically well-represented in the source documents.

\section{Discussion}
\label{sec:discussion}

We have documented a novel hybridized approach towards extractive summarization in this paper. This approach combines information from both statistical sources (keyword frequency) and semantic ones (using extracted entity relationships). In evaluation, we demonstrate that the resultant model  pleasantly combines strengths and ameliorates weaknesses of both varieties of summarization approaches. For instance, it provides superior coverage of document text topics than purely statistical methods, which sometimes focuses exclusively on the most prominent topic in the document. Unlike purely semantic methods, which may omit assessment of the relative importance of different topics in the document, to the detriment of the summary's coherence, our approach remains sensitive to these relative differences, resulting in coherent and readable summaries. 

It is also pertinent to note that our method is competitive against both classical baselines and state-of-the-art methods, since state-of-the-art methods have not shifted performance beyond the baseline significantly. We made some approximate relative comparisons between our approach and the more recent methods which were not a part of our eight summarizers that belonged to our pipeline of processing. Since the new approaches were not tested in our pipeline, the conditions in which they were reported are different and making a direct comparison between their reported ROUGE scores and ours is incorrect since the scores depend on many factors like the preprocessing steps, stemming, removal of stopwords, etc. In order to make a fair comparison between the performance of our approach and some new methods~\cite{mcdonald,litvaklastvanetik,gillickfavre}, we used TextRank scores as an intermediate basis to normalize the reported scores of the new approaches. This is made possible because the TextRank paper~\cite{textrank} reports different scores under different conditions (basic, stemmed, stemmed and no stopwords) which happen to be the conditions the new approaches were tested in. Thus, we would know how the new approaches fair with respect to TextRank relatively. On the other hand, we know how our approach fairs against TextRank. Thus, an indirect way of comparion is the closest we can get to make any comparison without having to actually run the new approaches in our pipeline. We also limited all the summaries to only 100 words per summary thus being as fair as possible. We have used the DUC 2002 dataset for comparison since all the other approaches have reported scores on the same dataset.

From this normalized relative comparison, we observed that the ROUGE-1 F score reported in McDonald, 2007 paper~\cite{mcdonald} over DUC 2002 dataset was 29.03\% lower than TextRank~\cite{textrank} for the same conditions after normalization. Litvak, Last, Vanetik, 2015 paper~\cite{litvaklastvanetik} reported ROUGE-1 recall scores for three approaches - their own approach (Gamp), McDonald, 2007~\cite{mcdonald} (McDonald) , and Gillick and Favre, 2009~\cite{gillickfavre} (Gillick). Gamp and McDonald had a respective increase of only 4.10\% and 3.39\% in ROUGE-1 recall while Gillick had a 0.93\% decrease in ROUGE-1 recall when compared to the recall scores published in the TextRank paper over DUC 2002 dataset for the same conditions and after normalization. We present a 7.93\% increase in ROUGE-1 F score and a 17.12\% increase in ROUGE-1 recall over TextRank for the same dataset and conditions, and after normalization. Thus, in addition to the five evaluated classical summarizers, our approach also works better than the more recent approaches.

The ROUGE results measure the summary size, overlap between the system and reference summary, and other statistical measures which, though extensive, are not holistic to know the quality of the summary. On the other hand, the survey evaluation would directly give what the ROUGE evaluation tries to capture i.e. to be as close to the human-like summary. Though it gives a fair amount of idea about the features of the summaries, the survey comes with its own biases until the size of participants is huge enough for proper aggregation of responses. Thus, reporting only ROUGE metrics or only survey results gives incomplete information about the summary. Most papers report only ROUGE results but it is important to consider both the ROUGE and survey results to come to a sound conclusion. Our holistic evaluation using both ROUGE and human evaluators permits us to conclude with significant confidence that our approach is competitive with the state-of-the-art, and ameliorates problems faced by purely frequency-based methods on the one hand, and purely topic- or graph-based methods on the other.

The datasets used had news reports which contain strong relationship between entities and our hypothesis is that our approach works better for such texts than for texts with weaker relationship between entities. On comparison with few classical methods and few recent state-of-the-art methods in extractive text summarization, we note that our approach captures more context and coverage of text, and ranks first on the DUC 2001 and 2002 datasets in terms of recall and F score for ROUGE measures. We have also shown how our approach captures more relevant text for summaries shorter than other baselines. In addition to this, our approach ranks the best in informativeness, coherence, and the ability to trick people into thinking that our summary was human-generated. Thus, our approach, using keyword-frequency to filter out less relevant text-entities before constructing entity relationship graphs, leads to selection of sentences that form overall better summaries. 

Finally we note that the present construction and evaluation is not yet optimized, so there is also considerable scope for improvement. For instance, the [Subject, Action, Object] format was obtained by using OpenIE and its native coreference resolver. A lot of post processing had to be done to improve the results of the OpenIE and remove noise. Any improvement in the way we get [Subject, Action, Object] links will help in getting better results.

Also, in our current implementation, the sentences were ranked in descending order based on the presence of important keywords in it and accordingly top sentences were selected. However, improvements in precision, and in turn F score, are likely if we normalize by sentence length i.e.,
\begin{equation}
    \label{ratio}
    \text{normalized score} = \frac{\text{total sentence score}}{\text{f(sentence length)}}
\end{equation}

This would, as we anticipate above, address our algorithm's current bias towards picking longer sentences, thus improving its precision vis-a-vis the ROUGE metric. Identifying a suitable normalization function $f$ for this task is an active focus of our future research.

\bibliographystyle{ACM-Reference-Format}
\bibliography{codscomad-bibliography}


\begin{thebibliography}{18}


\ifx \showCODEN    \undefined \def \showCODEN     #1{\unskip}     \fi
\ifx \showDOI      \undefined \def \showDOI       #1{#1}\fi
\ifx \showISBNx    \undefined \def \showISBNx     #1{\unskip}     \fi
\ifx \showISBNxiii \undefined \def \showISBNxiii  #1{\unskip}     \fi
\ifx \showISSN     \undefined \def \showISSN      #1{\unskip}     \fi
\ifx \showLCCN     \undefined \def \showLCCN      #1{\unskip}     \fi
\ifx \shownote     \undefined \def \shownote      #1{#1}          \fi
\ifx \showarticletitle \undefined \def \showarticletitle #1{#1}   \fi
\ifx \showURL      \undefined \def \showURL       {\relax}        \fi
\providecommand\bibfield[2]{#2}
\providecommand\bibinfo[2]{#2}
\providecommand\natexlab[1]{#1}
\providecommand\showeprint[2][]{arXiv:#2}

\bibitem[\protect\citeauthoryear{Angeli, Premkumar, and Manning}{Angeli
  et~al\mbox{.}}{2015}]%
        {angeli2015leveraging}
\bibfield{author}{\bibinfo{person}{Gabor Angeli}, \bibinfo{person}{Melvin
  Jose~Johnson Premkumar}, {and} \bibinfo{person}{Christopher~D. Manning}.}
  \bibinfo{year}{2015}\natexlab{}.
\newblock \showarticletitle{Leveraging linguistic structure for open domain
  information extraction}. In \bibinfo{booktitle}{\emph{Proceedings of the 53rd
  Annual Meeting of the Association for Computational Linguistics and the 7th
  International Joint Conference on Natural Language Processing (Volume 1: Long
  Papers)}}, Vol.~\bibinfo{volume}{1}. \bibinfo{pages}{344--354}.
\newblock


\bibitem[\protect\citeauthoryear{Baralis, Cagliero, Jabeen, Fiori, and
  Shah}{Baralis et~al\mbox{.}}{2013}]%
        {baralis2013multi}
\bibfield{author}{\bibinfo{person}{Elena Baralis}, \bibinfo{person}{Luca
  Cagliero}, \bibinfo{person}{Saima Jabeen}, \bibinfo{person}{Alessandro
  Fiori}, {and} \bibinfo{person}{Sajid Shah}.} \bibinfo{year}{2013}\natexlab{}.
\newblock \showarticletitle{Multi-document summarization based on the Yago
  ontology}.
\newblock \bibinfo{journal}{\emph{Expert Systems with Applications}}
  \bibinfo{volume}{40}, \bibinfo{number}{17} (\bibinfo{year}{2013}),
  \bibinfo{pages}{6976--6984}.
\newblock


\bibitem[\protect\citeauthoryear{Belica}{Belica}{2013}]%
        {sumy}
\bibfield{author}{\bibinfo{person}{Michal Belica}.}
  \bibinfo{year}{2013}\natexlab{}.
\newblock \bibinfo{booktitle}{\emph{Sumy Library - Python Package}}.
\newblock
\newblock
\shownote{{https://pypi.org/project/sumy/}.}


\bibitem[\protect\citeauthoryear{Brin and Page}{Brin and Page}{1998}]%
        {pagerank}
\bibfield{author}{\bibinfo{person}{Sergey Brin} {and} \bibinfo{person}{Lawrence
  Page}.} \bibinfo{year}{1998}\natexlab{}.
\newblock \showarticletitle{The anatomy of a large-scale hypertextual Web
  search engine}.
\newblock \bibinfo{journal}{\emph{Computer Networks and ISDN Systems}}
  \bibinfo{volume}{30} (\bibinfo{year}{1998}), \bibinfo{pages}{107--117}.
\newblock


\bibitem[\protect\citeauthoryear{DUC}{DUC}{2001}]%
        {duc2001}
\bibfield{author}{\bibinfo{person}{DUC}.} \bibinfo{year}{2001}\natexlab{}.
\newblock \bibinfo{booktitle}{\emph{Document understanding conference 2001}}.
\newblock
\newblock
\shownote{{http://www-nlpir.nist.gov/projects/duc/}.}


\bibitem[\protect\citeauthoryear{DUC}{DUC}{2002}]%
        {duc2002}
\bibfield{author}{\bibinfo{person}{DUC}.} \bibinfo{year}{2002}\natexlab{}.
\newblock \bibinfo{booktitle}{\emph{Document understanding conference 2002}}.
\newblock
\newblock
\shownote{{http://www-nlpir.nist.gov/projects/duc/}.}


\bibitem[\protect\citeauthoryear{Edmundson}{Edmundson}{1969}]%
        {edmundson}
\bibfield{author}{\bibinfo{person}{H.~P. Edmundson}.}
  \bibinfo{year}{1969}\natexlab{}.
\newblock \showarticletitle{New Methods in Automatic Extracting}.
\newblock \bibinfo{journal}{\emph{Journal of the Association for Computing
  Machinery}} \bibinfo{volume}{16}, \bibinfo{number}{2} (\bibinfo{year}{1969}),
  \bibinfo{pages}{264--285}.
\newblock


\bibitem[\protect\citeauthoryear{Erkan and Radev}{Erkan and Radev}{2004}]%
        {lexrank}
\bibfield{author}{\bibinfo{person}{G\"unes Erkan} {and}
  \bibinfo{person}{Dragomir~R. Radev}.} \bibinfo{year}{2004}\natexlab{}.
\newblock \showarticletitle{LexRank: Graph-based Lexical Centrality As Salience
  in Text Summarization}.
\newblock \bibinfo{journal}{\emph{Journal of Artificial Intelligence Research}}
  (\bibinfo{year}{2004}), \bibinfo{pages}{457--479}.
\newblock


\bibitem[\protect\citeauthoryear{Gillick and Favre}{Gillick and Favre}{2009}]%
        {gillickfavre}
\bibfield{author}{\bibinfo{person}{Dan Gillick} {and} \bibinfo{person}{Benoit
  Favre}.} \bibinfo{year}{2009}\natexlab{}.
\newblock \showarticletitle{A Scalable Global Model for Summarization}. In
  \bibinfo{booktitle}{\emph{Proceedings of the NAACL HLT Workshop on Integer
  Linear Programming for Natural Language Processing}}.
  \bibinfo{pages}{10--18}.
\newblock


\bibitem[\protect\citeauthoryear{Gong and Liu}{Gong and Liu}{2001}]%
        {lsa}
\bibfield{author}{\bibinfo{person}{Yihong Gong} {and} \bibinfo{person}{Xin
  Liu}.} \bibinfo{year}{2001}\natexlab{}.
\newblock \showarticletitle{Generic Text Summarization Using Relevance Measure
  and Latent Semantic Analysis}. In \bibinfo{booktitle}{\emph{Proceedings of
  the 24th Annual International ACM SIGIR Conference on Research and
  Development in Information Retrieval}}. \bibinfo{pages}{19--25}.
\newblock


\bibitem[\protect\citeauthoryear{Hennig, Umbrath, and Wetzker}{Hennig
  et~al\mbox{.}}{2008}]%
        {hennig2008ontology}
\bibfield{author}{\bibinfo{person}{Leonhard Hennig}, \bibinfo{person}{Winfried
  Umbrath}, {and} \bibinfo{person}{Robert Wetzker}.}
  \bibinfo{year}{2008}\natexlab{}.
\newblock \showarticletitle{An ontology-based approach to text summarization}.
  In \bibinfo{booktitle}{\emph{Proceedings of the 2008 IEEE/WIC/ACM
  International Conference on Web Intelligence and Intelligent Agent
  Technology-Volume 03}}. IEEE Computer Society, \bibinfo{pages}{291--294}.
\newblock


\bibitem[\protect\citeauthoryear{Lin}{Lin}{2004}]%
        {rougepaper}
\bibfield{author}{\bibinfo{person}{Chin-Yew Lin}.}
  \bibinfo{year}{2004}\natexlab{}.
\newblock \showarticletitle{ROUGE: A Package for Automatic Evaluation of
  Summaries}. In \bibinfo{booktitle}{\emph{Text Summarization Branches Out:
  Proceedings of the ACL-04 Workshop}}. \bibinfo{pages}{74--81}.
\newblock
\newblock
\shownote{Association for Computational Linguistics, Barcelona, Spain.}


\bibitem[\protect\citeauthoryear{Litvak, Last, and Vanetik}{Litvak
  et~al\mbox{.}}{2015}]%
        {litvaklastvanetik}
\bibfield{author}{\bibinfo{person}{Marina Litvak}, \bibinfo{person}{Mark Last},
  {and} \bibinfo{person}{Natalia Vanetik}.} \bibinfo{year}{2015}\natexlab{}.
\newblock \showarticletitle{Krimping texts for better summarization}. In
  \bibinfo{booktitle}{\emph{Proceedings of the 2015 Conference on Empirical
  Methods in Natural Language Processing}}. \bibinfo{pages}{1931--1935}.
\newblock
\newblock
\shownote{Lisbon, Portugal, September. Association for Computational
  Linguistics.}


\bibitem[\protect\citeauthoryear{Luhn}{Luhn}{1958}]%
        {luhn}
\bibfield{author}{\bibinfo{person}{Hans~Peter Luhn}.}
  \bibinfo{year}{1958}\natexlab{}.
\newblock \showarticletitle{The Automatic Creation of Literature Abstracts}.
\newblock \bibinfo{journal}{\emph{IBM Journal of Research Development}}
  \bibinfo{volume}{2} (\bibinfo{year}{1958}), \bibinfo{pages}{159--165}.
\newblock


\bibitem[\protect\citeauthoryear{McDonald}{McDonald}{2007}]%
        {mcdonald}
\bibfield{author}{\bibinfo{person}{Ryan McDonald}.}
  \bibinfo{year}{2007}\natexlab{}.
\newblock \showarticletitle{A study of global inference algorithms in
  multi-document summarization}. In \bibinfo{booktitle}{\emph{ECIR'07:
  Proceedings of the 29th European conference on IR research}}.
  \bibinfo{pages}{557--564}.
\newblock
\newblock
\shownote{Berlin, Heidelberg. Springer-Verlag.}


\bibitem[\protect\citeauthoryear{Mihalcea and Tarau}{Mihalcea and
  Tarau}{2004}]%
        {textrank}
\bibfield{author}{\bibinfo{person}{Rada Mihalcea} {and} \bibinfo{person}{Paul
  Tarau}.} \bibinfo{year}{2004}\natexlab{}.
\newblock \showarticletitle{Textrank: Bringing order into texts}. In
  \bibinfo{booktitle}{\emph{Proceedings of the 2004 Conference on Empirical
  Methods in Natural Language Processing. Association for Computational
  Linguistics}}. \bibinfo{pages}{404--411}.
\newblock


\bibitem[\protect\citeauthoryear{Rose, Engel, Cramer, and Cowley}{Rose
  et~al\mbox{.}}{2010}]%
        {rake}
\bibfield{author}{\bibinfo{person}{Stuart Rose}, \bibinfo{person}{Dave Engel},
  \bibinfo{person}{Nick Cramer}, {and} \bibinfo{person}{Wendy Cowley}.}
  \bibinfo{year}{2010}\natexlab{}.
\newblock \bibinfo{booktitle}{\emph{Automatic keyword extraction from
  individual documents}}.
\newblock
\newblock
\shownote{Text Mining: Applications and Theory, pages 1--20.}


\bibitem[\protect\citeauthoryear{Tomlin}{Tomlin}{1986}]%
        {tomlin1986basic}
\bibfield{author}{\bibinfo{person}{Russel~S. Tomlin}.}
  \bibinfo{year}{1986}\natexlab{}.
\newblock \bibinfo{booktitle}{\emph{Basic word order: functional principles.
  London: Croom Helm}}.
\newblock \bibinfo{publisher}{Coherence and grounding in discourse}.
\newblock


\end{thebibliography}

\end{document}